# Bayesian Networks from the Point of View of Chain Graphs*

Milan Studený[†]
Institute of Information Theory and Automation
Academy of Sciences of Czech Republic
Pod vodárenskou věží 4, 182 08 Prague, Czech Republic

## Abstract

The paper gives a few arguments in favour of use of chain graphs for description of probabilistic conditional independence structures. Every Bayesian network model can be equivalently introduced by means of a factorization formula with respect to chain graph which is Markov equivalent to the Bayesian network. A graphical characterization of such graphs is given. The class of equivalent graphs can be represented by a distinguished graph which is called the *largest chain graph*. The factorization formula with respect to the largest chain graph is a basis of a proposal how to represent the corresponding (discrete) probability distribution in a computer (i.e. 'parametrize' it). This way does not depend on the choice of a particular Bayesian network from the class of equivalent networks and seems to be the most efficient way from the point of view of memory demands.

A separation criterion for reading independences from a chain graph is formulated in a simpler way. It resembles the well-known $d$-separation criterion for Bayesian networks and can be implemented 'locally'.

## 1 INTRODUCTION

Two traditional approaches to description of probabilistic conditional independence structures use undirected graphs (Markov networks) and directed acyclic graphs (Bayesian networks) - see (Pearl 1988). Markov networks have *lines* (undirected edges) while Bayesian networks have *arrows* (directed edges). In middle eighties Lauritzen and Wermuth (1984) introduced the class of *chain graphs*, that is acyclic hybrid graphs having both lines and arrows. Since then many theoretical results analogous to the results concerning Bayesian networks were achieved - for an overview see (Studený 1996).

Chain graphs provide an elegant unifying point of view on Markov and Bayesian networks. However, researchers interested in graphical modelling of probabilistic structure may still have sensible objections against (wider acceptance of) chain graphs. Let us mention three of them.

1. The original way of introducing of the class of Markovian distributions with respect to a chain graph by means of the moralization criterion (Lauritzen 1989) seems to be too complex to be remembered immediately. It does not lead directly to an evident interpretation of chain graph models.

2. The first version of the separation criterion for chain graphs (Bouckaert and Studený 1995) seems even more complicated, especially in comparison with $d$-separation criterion for Bayesian networks. Moreover, it is not evident whether it can be implemented locally.

3. Many researchers believe that Bayesian networks provide a sufficiently general class of probabilistic models. Why one should use a more complex models for description of certain situations if one can describe it by a Bayesian network?

The aim of this paper is to respond to these three possible objections by adequate arguments in favour of chain graphs. In Section 2 some basic concepts are recalled. Then, in Section 3 we show that the class of Markovian distributions with respect to a chain graphs can be sometimes introduced very simply: by means of a factorization formula with respect to the chain graph. This holds in the case of a discrete strictly positive probability distribution and also in the case of a chain graph which is Markov equivalent to a Bayesian network (by equivalence is meant that the graphs de-scribe the same probabilistic model, that is they have the same class of Markovian distributions). The formula leads to a more natural explanation of meaning of a chain graph model: such a model is a result of combination of pieces of structural information obtained from different experts.

*This work was supported by the grants of Grant Agency of Czech Republic n. 201/98/0478, of Grant Agency of Academy of Sciences of Czech Republic n. K1075601 and n. A1075801 and of Ministry of Education of Czech Republic n. VS98008.

[†]The second affiliation is Laboratory of Intelligent Systems, University of Economics, Ekonomická 957, 14800 Prague, Czech Republic. E-mail: studeny@utia.cas.cz

In Section 4 a graphical characterization of chain graphs equivalent to Bayesian networks is given. Moreover, we propose to represent every equivalence class of Bayesian networks by the *largest chain graph* of the corresponding class of equivalent chain graphs. Note for explanation that the concept of largest chain graph is not identical with the concept of essential graph from (Andersson et. al. 1997a). Such a way of representation has certain advantage. First, one has a relatively simple graphical characterization of the largest chain graphs (equivalent to Bayesian networks) in graphical terms. Then every Bayesian network within the equivalence class can be obtained from the largest chain graph by directing its edges. Second, every chain graph within the corresponding equivalence class (including every mentioned Bayesian network) leads to a certain factorization formula which can be used to store Markovian distributions in memory of a computer. The formula with respect to the largest chain graph seems to lead to the most effective way from the point of view of memory demands (among formulas).

In Section 5 a simplified version of the separation criterion for chain graphs is presented. It resembles the well-known $d$-separation criterion for Bayesian networks very much. Moreover, we show that it can be implemented locally. Section 6 (Conclusions) summarizes content.

## 2 BASIC CONCEPTS

Throughout the paper $N$ will denote a non-empty finite set of *variables* which will be identified with nodes of graphs. For sake of brevity, juxtaposition $UV$ will denote the union $U \cup V$ of sets of variables $U, V \subset N$.

### 2.1 GRAPHS

A *hybrid graph* $G$ over $N$ (= the set of *nodes*) is specified by a set of two-element subsets of $N$, called *edges*, where for every edge $\{u, v\}$ just one of the following three cases occurs. Either it is a *line* between $u$ and $v$ (= undirected edge), denoted by $u - v$, or an *arrow* from $u$ to $v$ (= directed edge), denoted by $u \to v$, or an arrow from $v$ to $u$, denoted by $u \leftarrow v$. An *undirected graph* is a hybrid graph without arrows, a *directed graph* is a hybrid graph without lines. The *underlying graph* of $G$ is obtained from $G$ by replacing arrows with lines. Every set $\emptyset \neq A \subset N$ induces a subgraph over $A$, denoted by $G_A$, which has exactly those edges in $G$ which are subsets of $A$. *Components* of $G$ are obtained by removing all arrows of $G$ and taking the connectivity components of the remaining undirected graph.

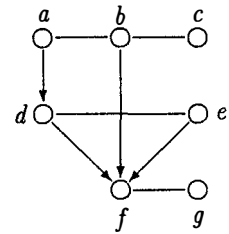

Figure 1: Example of a chain graph.

A *chain graph* is a hybrid graph whose components can be ordered into a sequence $C_1, \ldots, C_n$, $n \geq 1$ (called a *chain*) such that if $\{u, v\}$ is an edge with $u, v \in C_i$ then $u - v$, and if $\{u, v\}$ is an edge with $u \in C_i$, $v \in C_j$, $i < j$ then $u \to v$. For several other equivalent definitions see (Studený 1997). The reader can easily verify that chain graphs involve both undirected graphs and directed acyclic graphs. Figure 1 gives an example of a chain graph. The corresponding chain of components is $\{a, b, c\}$, $\{d, e\}$, $\{f, g\}$.

The set of *parents* of a set $A \subset N$ denoted by $pa_G(A)$ is $\{v \in N \setminus A \,;\, v \to u \text{ for some } u \in A\}$. The symbol of the graph $G$ can be omitted if it is clear from context. A sequence of distinct nodes $v_1, \ldots, v_k$, $k \geq 1$ such that $\{v_i, v_{i+1}\}$ is an edge in $G$ for every $i = 1, \ldots, k-1$ is called a *path* in $G$.

### 2.2 PROBABILITY

Given a collection of non-empty finite sets $\{\mathbf{X}_i; i \in N\}$, and $\emptyset \neq A \subset N$ the symbol $\mathbf{X}_A$ will denote $\prod_{i \in A} \mathbf{X}_i$. By convention $\mathbf{X}_\emptyset$ is a fixed singleton. Whenever $\mathbf{x} = [\mathbf{x}_i]_{i \in N} \in \mathbf{X}_N$ the symbol $\mathbf{x}_A$ will denote its projection $[\mathbf{x}_i]_{i \in A}$ to $\mathbf{X}_A$.

A *probability distribution* over $N$ is specified by a collection of non-empty finite sets $\{\mathbf{X}_i; i \in N\}$ and by a function

$$P: \mathbf{X}_N \to [0, 1] \quad \text{with} \quad \sum \{P(\mathbf{x}); \mathbf{x} \in \mathbf{X}_N\} = 1.$$

If $P(\mathbf{x}) > 0$ for all $\mathbf{x} \in \mathbf{X}_N$, then $P$ is called *strictly positive*. The *marginal distribution* of $P$ for $\emptyset \neq A \subset N$ is a probability distribution $P^A$ (over $A$) defined by:

$$P^A(\mathbf{y}) = \sum \{P(\mathbf{x}); \mathbf{x} \in \mathbf{X}_N, \mathbf{x}_A = \mathbf{y}\} \quad \text{for } \mathbf{y} \in \mathbf{X}_A.$$

We accept the convention $P^\emptyset(-) \equiv 1$. Having disjoint $A, B \subset N$, the *conditional probability* $P_{A|B}$ is a function on $\mathbf{X}_A \times \mathbf{X}_B$ defined by

$$P_{A|B}(\mathbf{x}_A | \mathbf{x}_B) = \frac{P^{AB}([\mathbf{x}_A, \mathbf{x}_B])}{P^B(\mathbf{x}_B)}$$

for $\mathbf{x}_A \in \mathbf{X}_A$, $\mathbf{x}_B \in \mathbf{X}_B$ in case $P^B(\mathbf{x}_B) > 0$. We accept the convention that $P_{A|B}(\mathbf{x}_A | \mathbf{x}_B) = 0$ whenever $P^B(\mathbf{x}_B) = 0$. Note that $P_{A|\emptyset} = P^A$.

Let us denote by $\mathcal{T}(N)$ the class of triplets $\langle A, B|C \rangle$ of disjoint subsets of $N$ whose first two components $A$

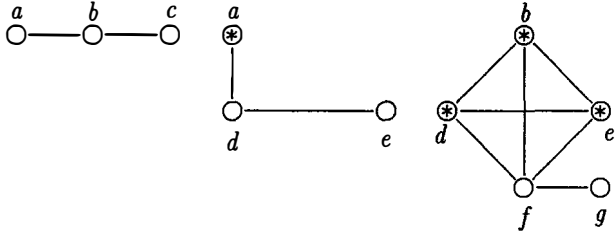

Figure 2: Closure graphs.

and $B$ are nonempty. Having $\langle A, B | C \rangle \in \mathcal{T}(N)$ and a probability distribution $P$ over $N$ we say that $A$ is *conditionally independent* of $B$ given $C$ with respect to $P$ and write $A \perp\!\!\!\perp B \mid C \ [P]$ if

$$P_{A|BC}(\mathbf{x}_A | [\mathbf{x}_B, \mathbf{x}_C]) = P_{A|C}(\mathbf{x}_A | \mathbf{x}_C)$$

for every $\mathbf{x}_A \in \mathbf{X}_A$, $\mathbf{x}_B \in \mathbf{X}_B$, $\mathbf{x}_C \in \mathbf{X}_C$.

## 3 FACTORIZATION FORMULA

The aim of this section is to show that probability distributions and chain graphs can be related quite simply by means of a factorization formula for chain graphs which has a natural intuitive interpretation.

A basic concept is factorization with respect to an undirected graph. Given an undirected graph $G$ over $N$, a set $A \subset N$ is complete if $u - v$ for every distinct $u, v \in A$. A *clique* of $G$ is a maximal complete set with respect to inclusion. A non-negative function $f$ defined on $\mathbf{X}_N$ is factorizable according to $G$ if there exists a collection of non-negative functions $\{\psi_K ; K \in \mathcal{K}\}$ where $\mathcal{K}$ the class of cliques of $G$ and $\psi_K$ is defined on $\mathbf{X}_K$ such that

$$f(\mathbf{x}) = \prod_{K \in \mathcal{K}} \psi_K(\mathbf{x}_K) \quad \text{for } \mathbf{x} \in \mathbf{X}_N.$$

### 3.1 FACTORIZABLE DISTRIBUTIONS

Let $G$ be a chain graph over $N$ and $C$ a component of $G$. Then by the *closure graph* for $C$ we understood the undirected graph $H(C)$ over $C \cup pa_G(C)$ obtained in this way:

- take lines and arrows in $G$ (directed) to nodes of $C$ and drop direction (change arrows into lines),
- connect by a line every pair of distinct nodes of $pa_G(C)$.

Equivalently, make $pa_G(C)$ complete in the underlying graph of $G_{C \cup pa(C)}$. Figure 2 shows the closure graphs for components of the graph from Figure 1. Nodes of $pa(C)$ are marked by an asterisk.

A probability distribution $P$ over $N$ is *factorizable* with respect to a chain graph $G$ over $N$ if the following two conditions hold:

(a) component-wise factorization, that is

$$P(\mathbf{x}) = \prod_{C \in \mathcal{C}} P_{C|pa(C)}(\mathbf{x}_C | \mathbf{x}_{pa(C)}) \quad \text{for } \mathbf{x} \in \mathbf{X}_N$$

where $\mathcal{C}$ is the class of components of $G$.

(b) clique-wise factorization, that is for every component $C$ of $G$ the conditional distribution $P_{C|pa(C)}$ is factorizable according to the closure graph $H(C)$.

The condition (b) can be equivalently expressed as the requirement that the marginal distribution $P^{C \cup pa(C)}$ is factorizable according to $H(C)$, or that $P_{C|pa(C)}$ is computed from a distribution $Q$ over $C \cup pa_G(C)$ which is factorizable according to $H(C)$.

EXAMPLE 3.1 Let $G$ be the graph from Figure 1. A probability distribution $P$ over $\{a, b, c, d, e, f, g\}$ is factorizable with respect to $G$ iff

$$P = P^{abc} \cdot P_{de|a} \cdot P_{fg|bde},$$

and moreover

$$P^{abc} = \psi_{ab} \cdot \psi_{bc}, \quad P_{de|a} = \psi_{ad} \cdot \psi_{de}, \quad P_{fg|bde} = \psi_{bdef} \cdot \psi_{fg}$$

where $\psi$'s are arbitrary respective non-negative functions.

In the preceding example, the elementary factors are not specified. Indeed, one has freedom in their choice in general. In the next section we will mention more specific factorizations in which the factors are conditional (marginal) probabilities computed from $P$.

Such a two-degree factorization formula does not look natural at first sight. However, it has very good intuitive interpretation. Let us consider a situation when our probabilistic model is composed from pieces of (structural) information obtained from different experts. Thus, each expert has his/her exclusive non-empty area of competence; the areas are disjoint and cover together the whole set of factors $N$. The expert is supposed not only to give information about structural relationships within his/her area of competence but also indicate outside factors which influence the factors within the area of competence. To prevent discrepancy between experts we order the areas of competence into a sequence $C_1, \ldots, C_n$, $n \geq 1$ and ask every expert to indicate the influential factors from preceding areas only. That is, the $i$-th expert can 'provide' information about the conditional probability $P_{C_i|C_1 \cup \ldots \cup C_{i-1}}$ $(i = 1, \ldots, n)$ in the form of $P_{C_i|pa(C_i)}$ where $pa(C_i) \subset C_1 \cup \ldots \cup C_{i-1}$ is the influence area determined by the expert. However, we suppose that the expert can provide more detailed information about the structure of $P_{C_i|pa(C_i)}$. Thus, we want him/her to indicate how it factorizes, that is to provide the corresponding undirected graph $H(C_i)$ over $C_i \cup pa(C_i)$. Since the expert has $C_i$ as the limited area of competence, he/she is not entitled to evidence relationships within $pa(C_i)$. Thus, it is natural to suppose that $pa(C_i)$ is a complete in $H(C_i)$ (this is mathematical representation of 'missing' structural information). For simplicity we suppose that $C_i$ is a connected subset in each $H(C_i)$. Altogether, $P$ is factorizable with respect to a certain chain graph over $N$ having $C_1, \ldots, C_n$ as components. Indeed, for every $C_i$, create 'local' hybrid graph $G(C_i)$ from $H(C_i)$: remove edges within $pa(C_i)$ and direct edges from $pa(C_i)$ to $C_i$. Then compose $G$ of these local hybrid graphs.

Let us mention a special case of Bayesian networks. In this case components are singletons and $P$ is factorizable with respect to an acyclic directed graph $G$ iff $P = \prod_{i \in N} P_{i|pa(i)}$. This well-known formula than can be interpreted as above: this time the $i$-th expert has $C_i = \{i\}$ as his/her area of competence and his/her only role is to indicate $pa(C_i)$ ($H(C_i)$ is complete!).

### 3.2 MARKOVIAN DISTRIBUTIONS

Let us recall the original way how probabilistic structure was ascribed to a chain graph (Lauritzen 1989), (Frydenberg 1990). Supposing $G$ is a chain graph over $N$, its *moral graph* is an undirected graph over $N$ in which $\{u, v\}$ is an edge iff $\{u, v\}$ is an edge in $G$ or $u, v \in pa_G(C)$, $u \neq v$ for a component $C$ of $G$. For example, the closure graph $H(C)$ mentioned in 3.1 is nothing but the moral graph of $G_{C \cup pa(C)}$. Having a set of nodes $A \subset N$ the symbol $an_G(A)$ denotes the set of *ancestors* of $A$, that is the set of those nodes $v \in N$ that there exists a path $v = w_1, \ldots, w_k = u$, $k \geq 1$ in $G$ from $v$ to a node $u \in A$ such that $w_i \to w_{i+1}$ or $w_i - w_{i+1}$ for $1 \leq i \leq k-1$. Note that $A \subset an_G(A)$. A triplet $\langle A, B|C \rangle \in \mathcal{T}(N)$ is *represented* in a chain graph $G$ (according to the *moralization criterion*) if every path in the moral graph of $G_{an(ABC)}$ from a node of $A$ to a node of $B$ contains a node of $C$.

A probability distribution $P$ over $N$ is *Markovian* with respect to a chain graph $G$ over $N$ if $A \perp\!\!\!\perp B \mid C\,[P]$ for every triplet $\langle A, B|C \rangle \in \mathcal{T}(N)$ represented in $G$.

LEMMA 3.1 *Every probability distribution factorizable with respect to a chain graph $G$ is Markovian with respect to $G$.*

**Proof:** Suppose that $P$ is factorizable with respect to $G$, and $\langle A, B|C \rangle \in \mathcal{T}(N)$ is represented in $G$ according to the moralization criterion. One can find a chain of components $C_1, \ldots, C_n$, $n \geq 1$ such that $an(ABC) = C_1 \cup \ldots \cup C_m$ for some $1 \leq m \leq n$. We leave it to the reader to verify using this fact that $P^{an(ABC)}$ is factorizable with respect to $G_{an(ABC)}$. Hence, $P^{an(ABC)}$ is factorizable according to the moral graph of $G_{an(ABC)}$. Therefore $P^{an(ABC)}$ is Markovian with respect to the moral graph of $G_{an(ABC)}$ - see (Lauritzen *et. al.* 1990). Hence $A \perp\!\!\!\perp B \mid C\,[P]$. □

Thus, the factorization property implies Markovness. The converse holds often, too. Frydenberg (1990) showed that Markovness implies factorization in case of strictly positive distributions. In sequel we show that both conditions are also equivalent for certain special chain graphs. Thus, the class of Markovian distributions can be often introduced very simply by means of the factorization formula.

## 4 MARKOV EQUIVALENCE

We say that two chain graphs over $N$ are *Markov equivalent* if their classes of Markovian distributions coincide. Let us recall Frydenberg's (1990) graphical characterization of Markov equivalence which generalizes an analogous result for Bayesian networks (Verma Pearl 1991).

A *complex* in a chain graph $G$ is a special induced subgraph of $G$, namely a path $v_1, \ldots, v_k$, $k \geq 3$, such that $v_1 \to v_2$, $v_i - v_{i+1}$ for $i = 2, \ldots, k-2$, $v_{k-1} \leftarrow v_k$ in $G$, and no additional edges between nodes of $\{v_1, \ldots, v_k\}$ exist in $G$. For example, the only complexes in the graph from Figure 1 are $b \to f \leftarrow d$ and $b \to f \leftarrow e$.

THEOREM 4.1 *Two chain graphs over $N$ are Markov equivalent iff they have the same underlying graph and the same complexes.*

### 4.1 CHAIN GRAPHS MARKOV EQUIVALENT TO BAYESIAN NETWORKS

An undirected graph $G$ is called *decomposable* if the collection of its cliques can be ordered into a sequence $K_1, \ldots, K_k$, $k \geq 1$ satisfying the *running intersection property*, that is

$$\forall i > 2 \ \exists j < i \quad K_i \cap (\bigcup_{l < i} K_l) \subset K_j \,.$$

Then, for every clique $K$ of $G$, one can find such an ordering which starts by $K$. For the proof of this fact and further equivalent definitions of a decomposable graph see (Lauritzen 1996). In (Andersson *et. al.* 1997b) the following result is shown.

THEOREM 4.2 *A chain graph $G$ is Markov equivalent to an acyclic directed graph (Bayesian network) iff for every component $C$ of $G$ the closure graph $H(C)$ is decomposable.*

An important fact is that a probability distribution $Q$ factorizable according to a decomposable undirected graph can be factorized in such a way that the factors are conditional probabilities (computed from $Q$). In fact, there exists a distinguished formula for $Q$ in terms of its marginals.

LEMMA 4.1 *Let $Q$ be a probability distribution factorizable according to a decomposable undirected graph $G$, and $K_1, \ldots, K_k$, $k \geq 1$ is an ordering of its cliques satisfying the running intersection property. Then*

$$Q = \prod_{i=1}^{k} \frac{Q^{K_i}}{Q^{K_i \cap (\bigcup_{l < i} K_l)}} \qquad (1)$$

$$Q = \prod_{i=1}^{k} Q_{K_i \setminus (\bigcup_{l<i} K_l) \mid K_i \cap (\bigcup_{l<i} K_l)} \quad (2)$$

**Proof:** One can use induction according to the number of cliques $k$. The asumption implies that $Q = \psi_L \cdot \psi_M$ where $L = \bigcup_{j<k} K_j$, $M = K_k$. We leave it to the reader to show that $Q = (Q^L \cdot Q^M)/Q^{L \cap M}$. Since $Q^L$ is factorizable according to $G_L$ (which is decomposable) one can apply the induction assumption to $Q^L$. □

The sets $S_i = K_i \cap (\bigcup_{l<i} K_l)$ determined by the chosen sequence $K_1, \ldots, K_k$ are called *separators* of the sequence (Lauritzen 1996). It may happen that $S_i = S_j$ for $i \neq j$. However, the set of separators and the number of their occurences in $S_1, \ldots, S_k$ does not depend on the choice of the sequence $K_1, \ldots, K_k$ satisfying the running intersection property - see for example Lemma 2.18 in (Studený 1992). In particular, the expression (1) for $Q$ in Lemma 4.1 does not depend on the choice of $K_1, \ldots, K_k$.

Thus, in case of a chain graph $G$ which equivalent to a Bayesian network the factorization formula from Section 3.1 can be made more specific. Indeed, for each component $C$ of $G$ one can choose a clique of the closure graph $H(C)$ containing $pa_G(C)$. Then, one can find an ordering $K_1, \ldots, K_{k(C)}$, $k(C) \geq 1$ of the cliques of $H(C)$ satisfying the running intersection property starting by the chosen clique and apply Lemma 4.1 to $Q = P^{C \cup pa(C)}$. After that, one can write by (2)

$$P_{C \mid pa(C)} = \frac{P^{K_1}}{P^{pa(C)}} \cdot \prod_{i=2}^{k(C)} P_{K_i \setminus S_i \mid S_i}$$

where $S_i$ are the corresponding separators. This leads to the following global factorization:

$$P = \prod_{C \in \mathcal{C}} \left( P_{K_1 \setminus pa(C) \mid pa(C)} \cdot \prod_{i=2}^{k(C)} P_{K_i \setminus S_i \mid S_i} \right).$$

Note that in case $G$ is directly an acyclic directed graph it collapses to the classical formula $P = \prod_{i \in N} P_{i \mid pa(i)}$.

EXAMPLE 4.1 To illustrate it let us continue with Example 3.1. Each closure graph from Figure 2 is decomposable. Both cliques of the component $\{a, b, c\}$ contain $pa(\{a, b, c\}) = \emptyset$. Hence, there are two possible ways of writing $P^{abc}$, that is either $P^{abc} = P^{ab} \cdot P_{c \mid b}$  or  $P^{abc} = P^{bc} \cdot P_{a \mid b}$.

However, the required ordering of cliques is unique for the other components:

$$P_{de \mid a} = P_{d \mid a} \cdot P_{e \mid d}, \quad P_{fg \mid bde} = P_{f \mid bde} \cdot P_{g \mid f}.$$

This leads to two factorization formulas for $P$ having conditional probabilities as basic factors. Nevertheless, one can also write it using (1) as a 'ratio of marginals' which leads to the following formula:

$$P = \frac{P^{ab} \cdot P^{bc}}{P^b} \cdot \frac{P^{ad} \cdot P^{de}}{P^a \cdot P^d} \cdot \frac{P^{bdef} \cdot P^{fg}}{P^{bde} \cdot P^f}.$$

As explained in the following remark one can obtain such an 'unique' formula for every chain graph which is Markov equivalent to a Bayesian network.

*Remark* In the formula before Example 4.1 $P$ is expressed as a product of conditional probabilities. The overall number of factors is the number of cliques in all closure graphs for components. However, as mentioned in Example 4.1, the formula may depend on the choice of orderings of cliques. Let us mention two ways how to avoid seeming ambiguity.

First, one can use the first expression (1) from Lemma 4.1 for $P^{C \cup pa(C)}$ which leads to the formula

$$P = \prod_{C \in \mathcal{C}} \frac{\prod_{K \in \mathcal{K}(C)} P^K}{P^{pa(C)} \cdot \prod_{S \in \mathcal{S}(C)} (P^S)^{m(S)}},$$

where $\mathcal{K}(C)$ is the class of cliques of $H(C)$, $\mathcal{S}(C)$ is the class of separators of $H(C)$, and $m(S)$ denotes the number of occurences of a separator $S$. Here, the elementary factors are only marginals of $P$. Seemingly, the number of factors both in numerator and denominator is the overall number of cliques. However, some factors can cancel out. I have some reasons to believe that this form of expression for $P$ even does not depend on the choice of the graph from the class of Markov equivalent chain graphs.

The second way is the formula which has elementary factors in the form of conditional probabilities $P_{A \setminus pa(C) \mid A \cap pa(C)}$ where $A$ is either a clique or a separator of $H(C)$ for a component $C$. Indeed, since $pa(C) \subset K_1$ one has $K_i \cap pa(C) = S_i \cap pa(C)$ for $i \geq 2$ and therefore

$$P = \prod_{C \in \mathcal{C}} \frac{\prod_{K \in \mathcal{K}(C)} P_{K \setminus pa(C) \mid K \cap pa(C)}}{\prod_{S \in \mathcal{S}(C)} (P_{S \setminus pa(C) \mid S \cap pa(C)})^{m(S)}}.$$

PROPOSITION 4.1 *Let $G$ be a chain graph which is Markov equivalent to a Bayesian network. Then every Markovian distribution with respect to $G$ is factorizable with respect to $G$.*

**Proof:** Let us fix a chain of components $C_1, \ldots, C_n$, $n \geq 1$ of $G$. Then use Theorem 4.2 and fix a respective ordering $K_1, \ldots, K_{k(C)}$, $k(C) \geq 1$ of cliques of $H(C)$ for every component $C$ (see above). Construct a sequence of all nodes of $N$ in the following way: take components in their order and within each component consider the blocks $B_1 = K_1 \setminus pa(C)$, $B_i = K_i \setminus \bigcup_{j<i} B_j$ for $i \geq 2$, in their order (the order within those blocks is immaterial). This ordering is consonant with orientation of arrows in $G$. One can direct all edges of $G$ according to this ordering and obtain an acyclic directed graph $D$ which is Markov equivalent to $G$ by Theorem 4.1. Thus, every distribution $P$ which is Markovian with respect to $G$ is Markovian with respect to $D$. By Theorem 1 in (Lauritzen et. al. 1990) $P$ is factorizable with respect to $D$, that is $P = \prod_{i \in N} P^{i \cup pa_D(i)}/P^{pa_D(i)}$. Hence, the reader can verify that $P$ is factorizable with respect to $G$. □

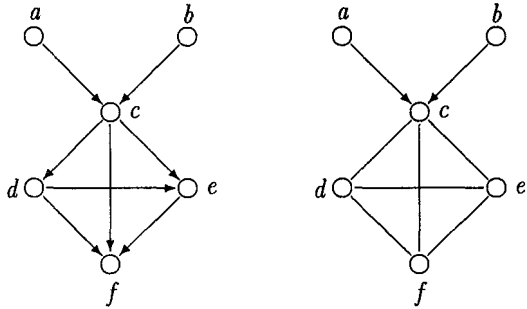

Figure 3: Example of a Bayesian network and the corresponding largest chain graph.

### 4.2 LARGEST CHAIN GRAPH

Supposing $G$ and $H$ are Markov equivalent chain graphs over $N$ we say that $G$ is *larger* than $H$ and write $H \prec G$ if every arrow in $G$ is an arrow in $H$ with the same orientation. Informally, $G$ has 'more' lines than $H$. Frydenberg (1990) showed that within each Markov equivalence class $\mathcal{G}$ of chain graph there exists a chain graph $G_\infty \in \mathcal{G}$ which is larger than every other $G \in \mathcal{G}$. The graph $G_\infty$ is then called the *largest chain graph* of $\mathcal{G}$. Let us recall a graphical characterization of the largest chain graphs from (Volf and Studený 1998).

An arrow $u \to v$ covers an arrow $x \to y$ in a chain graph $G$ if $u$ is an ancestor of $x$ and $y$ is an ancestor of $v$ in $G$. An arrow $u \to v$ is *protected* in $G$ if it covers an arrow which belongs to a complex in $G$. In particular, every complex arrow is protected.

THEOREM 4.3 *A chain graph is the largest chain graph (of a Markov equivalence class of chain graphs) iff every its arrow is protected.*

In fact, $G_\infty$ has an arrow $u \to v$ iff $u \to v$ is shared by all graphs from $\mathcal{G}$. By the preceding result, these shared arrows can be identified as protected arrows in every $G \in \mathcal{G}$.

EXAMPLE 4.2 Let us consider the acyclic directed graph $D$ in the left picture in Figure 3. The only protected arrows in $D$ are the arrows of the complex $a \to c \leftarrow b$. Thus, the corresponding largest chain graph $G_\infty$ is in the right picture of Figure 3. The factorization formula with respect to $D$ is

$$P = P^a \cdot P^b \cdot P_{c|ab} \cdot P_{d|c} \cdot P_{e|cd} \cdot P_{f|cde} .$$

Note that each Markov equivalent Bayesian network has the arrows $c \to d$, $c \to e$ and $c \to f$. Thus, Markov equivalent Bayesian networks differ only in permutation of nodes $d, e, f$. In particular, all equivalent Bayesian networks induce the same type of the factorization formula. On the other hand, the factorization formula with respect to $G_\infty$ is different:

$$P = P^a \cdot P^b \cdot P_{c|ab} \cdot P_{def|c} .$$

How it is related to Bayesian networks? Let $\mathcal{D}$ be a class of Markov equivalent Bayesian networks. Well, every $D \in \mathcal{D}$ is only an auxiliary tool to describe the (common) underlying probabilistic structure (induced by each $D \in \mathcal{D}$) which is the crucial concept. Is it possible to represent the structure by a distinguished graph which reflects (only) the substantial features of the structure? This problem has not an appropriate solution within the framework of acyclic directed graphs. All equivalent Bayesian networks from Example 4.2 are equally entitled to represent the corresponding probabilistic structure but none of them expresses exchangeability of $d, e, f$. However, perhaps a solution can be found in a wider class of graphs. Let us consider the class $\mathcal{G}$ of all chain graphs which are Markov equivalent to the graphs in $\mathcal{D}$. Of course, $\mathcal{G}$ is wider than $\mathcal{D}$ but the point is that the graphs from $\mathcal{G}$ describe the same probabilistic structure as the graphs from $\mathcal{D}$!

We propose to represent the structure by the largest chain graph $G_\infty$ of $\mathcal{G}$. Note that Theorem 4.3 together with Theorem 4.2 gives a graphical characterization of the largest chain graphs equivalent Bayesian networks. This approach has several advantages. First, the choice of the representative is made on basis of a 'fair' mathematical criterion - the maximal number of undirected edges. Second, every Bayesian network $D \in \mathcal{D}$ can be obtained from $G_\infty$ by directing all its lines. The point of view of chain graphs gives a better insight into the class of equivalent graphs. While $G_\infty$ is the maximal graph with respect to $\prec$ within $\mathcal{G}$ Bayesian networks are the minimal graphs. Perhaps the task of checking equivalence of two Bayesian networks $D_1$ and $D_2$ looks more transparent now. They are equivalent iff their corresponding largest chain graphs coincide. Instead of 'converting' arrows in $D_1$ to obtain $D_2$ one can apply an algorithm transforming both into the corresponding largest chain graph (Studený 1997). Third, the factorization formula with respect to the largest chain graph mentioned in 4.1 offers a promising method how to represent Markovian distributions in memory of a computer. Let us return to Example 4.2. The difference between respective formulas is that the term $P_{def|c}$ from the formula induced by $G_\infty$ is in the formula induced by $D$ formally disintegrated:

$$P_{def|c} = P_{d|c} \cdot P_{e|cd} \cdot P_{f|cde} .$$

However, the memory demands for $P_{def|c}$ and for $P_{f|cde}$ are the same: in both cases one needs $|\mathbf{X}_{cdef}|$ numbers.[1] Thus, the terms $P_{d|c}$ and $P_{e|cd}$ only raise memory demands. I think that this holds in general.

---

[1] The reader can object that only $|\mathbf{X}_B| \cdot (|\mathbf{X}_A| - 1)$ values suffices to represent $P_{A|B}$ since the values of a probability distribution sum to 1. In fact, this 'actual algebraic dimension' of $P_{def|c}$ and one of $P_{d|c} \cdot P_{e|cd} \cdot P_{f|cde}$ are the same, namely $|\mathbf{X}_c| \cdot (|\mathbf{X}_{def}| - 1)$. It does not depend on the choice of the graph at all, I guess. However, this extremely frugal way of 'parametrization' almost surely leads to other computational difficulties. I doubt whether this is

The number of factors in the factorization formula from 4.1 (the formula induced by a Bayesian network is a special case) is the number of cliques of all closure graphs. One can expect less number of these cliques in a graph with less number of components. And the minimal number of components within $\mathcal{G}$ has the largest chain graph of $\mathcal{G}$! An intuitive hope that the above mentioned approach could be computationally feasible is based on Theorem 4.2. Distributions complying with decomposable models can be represented in form of a junction tree. This is a basis of a local computation algorithm based on 'probability propagation' - see (Lauritzen 1996). Perhaps one can somehow to utilize the fact that in the formula from 4.1 every $P_{C|pa(C)}$ (or equivalently $P^{C \cup pa(C)}$) complies with 'local' decomposable model.

*Remark* Andersson et. al. (1997a) proposed to represent every class $\mathcal{D}$ of Markov equivalent Bayesian networks by the *essential* graph of $\mathcal{D}$ which has the same underlying graph as every $D \in \mathcal{D}$ and only those arrows which are shared by all graphs in $\mathcal{D}$. This is always a chain graph. Example 4.2 shows that the corresponding Markov equivalence class of chain graphs $\mathcal{G}$ is wider than $\mathcal{D}$ and therefore the essential graph does not coincide with the corresponding largest chain graph, in general.

## 5  SEPARATION CRITERION

The aim of this section is to formulate a separation criterion for chain graphs in such a way that its affinity with $d$-separation is evident. Moreover, we indicate how to implement it locally.

### 5.1  SUPERACTIVE ROUTES

A *route* in a chain graph $G$ is a sequence of nodes $\rho : v_1, \ldots, v_k$, $k \geq 1$ such that $\{v_i, v_{i+1}\}$ is an edge of $G$ for every $i = 1, \ldots, k - 1$. The difference from the concept of path is that nodes in a path are distinct, but the nodes in a route can be repeated! A *section* of $\rho$ is its maximal undirected subroute $v_i - \ldots - v_j$, $1 \leq i \leq j \leq k$. If $1 < i \leq j < k$, $v_{i-1} \rightarrow v_i$ and $v_j \leftarrow v_{j+1}$, then it is called a *head-to-head* section. In case of a Bayesian network sections are just single nodes. Suppose that $C$ is a set of nodes of a chain graph $G$. We say that $\rho$ is *superactive* with respect to $C$ iff

- every head-to-head section of $\rho$ has a node of $C$,
- every other section of $\rho$ is outside $C$.

A triplet $\langle A, B|C \rangle \in \mathcal{T}(N)$ is represented in $G$ (according to the *separation criterion*) if there is no route in $G$ from a node in $A$ to a node in $B$ which is superactive with respect to $C$.

---

an effective way of internal representation of distributions in a computer.

What does it mean in case of Bayesian networks? A route $\rho$ in an acyclic directed graph $D$ is superactive iff

$$u \in C \Leftrightarrow u \text{ is a head-to-head node}$$

for every node $u$ of $\rho$. This condition is stronger than the concept of active route (Pearl 1988). In an active route $\sigma$ non-head-to-head nodes are outside $C$ and head-to-head nodes have descendants in $C$. But whenever $u$ is a head-to-head node in $\sigma$ outside $C$, then there exists a path $u = w_1 \rightarrow \ldots \rightarrow w_k \in C$, $k \geq 2$ with $\{w_1, \ldots, w_{k-1}\} \cap C = \emptyset$ and $\sigma$ can be patched by $w_1 \rightarrow \ldots \rightarrow w_k \leftarrow \ldots \leftarrow w_1$. Thus, any active route can be modified into a superactive route. Therefore, one can formulate $d$-separation criterion in the following way. A triplet $\langle A, B|C \rangle \in \mathcal{T}(N)$ is represented in $D$ if there is no route between $A$ and $B$ which is superactive with respect to $C$. I think that this formulation of $d$-separation is even simpler than the original one. The surpising simplicity is due to the fact we consider the class of all routes which may be infinite. Every active route can be shortened to an active path but this is not true for superactive routes. In either case, it is evident that the above mentioned separation criterion for chain graphs generalizes (a simplified version of) $d$-separation.

Note that the original $c$-separation criterion (= separation criterion for chain graph) was more complicated. It was formulated for certain finite class of routes, called trails, and an auxiliary concept of slide was necessary. The proof of the following lemma is analogous to the proof of Proposition 3 from (Lauritzen et. al. 1990). We omit it for page limitation.

LEMMA 5.1 *Let $G$ be a chain graph over $N$. Then $\langle A, B|C \rangle \in \mathcal{T}(N)$ is represented in $G$ according to the separation criterion iff it is represented in $G$ according to the moralization criterion.*

### 5.2  ALGORITHM

Potentially infinite number of routes in a chain graph may cause doubts whether $c$-separation can be implemented on a computer. To show that is possible we propose an algorithm which by 'local propagation' indicated the nodes connected by a superactive route.

**Input** A chain graph $G$ over $N$, $A, C \subset N$ disjoint.

Since the class $\mathcal{I}$ of triplets represented in $G$ satisfies the following 'composition property' (Studený and Bouckaert 1998):

$$\langle A, B_1|C \rangle, \langle A, B_2|C \rangle \in \mathcal{I} \;\Rightarrow\; \langle A, B_1 \cup B_2|C \rangle \in \mathcal{I},$$

there exists unique maximal $B \subset N \setminus AC$ (possibly empty) such that $\langle A, B|C \rangle$ is represented in $G$.

**Output** Maximal $B \subset N \setminus AC$ such that $\langle A, B|C \rangle$ is represented in $G$ according to the separation criterion.

**Initiation** Four sets $U, V, W, Z \subset N$ will be modified dynamically by the algorithm. Put $U = A$, $V = W = Z = \emptyset$.

**Inference rules** This will be done by the use of the following 'propagation' rules applicable to edges $\{u,v\}$ in $G$ (one of them is applied to nodes $u \in N$ only).

1. $u \in U$, $u - v$, $v \notin C \Rightarrow v \in U$,
2. $u \in U$, $u \leftarrow v$, $v \notin C \Rightarrow v \in U$,
3. $u \in UV$, $u \to v$, $v \notin C \Rightarrow v \in V$,
4. $u \in V$, $u - v$, $v \notin C \Rightarrow v \in V$,
5. $u \in UV$, $u \to v \Rightarrow v \in W$,
6. $u \in W$, $u - v \Rightarrow v \in W$,
7. $u \in W$, $u \in C \Rightarrow v \in Z$,
8. $u \in Z$, $u - v \Rightarrow v \in Z$,
9. $u \in Z$, $u \leftarrow v$, $v \notin C \Rightarrow v \in U$.

**Stopping rule** The algorithm will end when preceding rules cannot enlarge the sets $U,V,W,Z$. Then put $B = N \setminus UVC$.

LEMMA 5.2 *The algorithm above indicates as $UV$ the nodes $v \in N$ such that there exists a route in $G$ from $u \in A$ to $v$ which is superactive with respect to $C$.*

**Proof:** Let us ascribe meaning to sets $U, V, W, Z$. The set $V$ contains those $v \in N$ such that there exists a superactive route from $u \in A$ to $v$ whose last section has the form $w_i \to w_{i+1} - \ldots - w_{i+k} = v$, $k \geq 1$. The set $U$ involves those $v \in N$ such that there exists a superactive route from $u \in A$ to $v$ whose last section has not such a form. The set $W$ contains those $v \in N$ that there exists $u \in UV$ and a route $u = t_0 \to t_1 - \ldots - t_r = v$, $r \geq 1$ in $G$. And $Z$ denotes $v \in N$ such that there exists a route $u = t_0 \to t_1 - \ldots - t_r = v$, $r \geq 1$ in $G$ with $u \in UV$ and $\{t_1,\ldots,t_r\} \cap C \neq \emptyset$. We leave it to the reader to check that the rules above have to indicate gradually all nodes of $U,V,W,Z$. □

## 6  CONCLUSIONS

Let us summarize the paper. Section 3 responds to an objection that the way of introducing the class of Markovian distributions for chain graphs is too complex. The factorization formula is quite simple, has reasonable interpretation, and fits in case of chain graphs equivalent to Bayesian networks (Lemma 3.1, Proposition 4.1). Section 4 tries to show that chain graphs can be useful even in situations which can be described by Bayesian networks. We propose to represent the class of equivalent Bayesian networks by the corresponding largest chain graph and argue that it leads to an effective way of computer representation of Markovian distributions. Section 5 responds to an objection that the separation criterion for reading independences from a chain graph is too complicated in comparison with $d$-separation. Much simpler version of $c$-separation is presented. Moreover, we propose a method how to implement it in such a way that in each step only neighbor nodes are consulted.

**Acknowledgments**

I would like to express my thanks to anomymous reviewers for their comments.